\documentclass[11pt,a4paper]{article}
\usepackage[hyperref]{acl2021}
\usepackage{pdflscape}
\usepackage{array}
\usepackage{footnote}
\usepackage[justification=centering]{caption}
\usepackage{afterpage}
\usepackage{tabularx} 
\usepackage{capt-of}
\usepackage{lipsum}
\usepackage{times}
\usepackage{latexsym}

\usepackage{url}
\newcommand{\tablefont}{\fontsize{3mm}{3mm}\selectfont}
\usepackage{tabularx}
\usepackage{graphicx}
\usepackage{adjustbox}
\usepackage{ntheorem}
\theoremseparator{:}

\usepackage{capt-of}
\usepackage{lipsum}
\usepackage{microtype}

\aclfinalcopy 

\title{A comparative analysis of embedding models for patent similarity}

\author{
  Grazia Sveva Ascione $^1$, Valerio Sterzi $^1$ \\
 $^1$ Bordeaux School of Economics (Université de Bordeaux) \\
\texttt{\{grazia.ascione, valerio.sterzi\}@u-bordeaux.fr} \\
}
\date{}

\begin{document}
\maketitle
\begin{abstract}
This paper makes two contributions to the field of text-based patent similarity. First, it compares the performance of different kinds of patent-specific pretrained embedding models, namely \textit{static} word embeddings (such as word2vec and doc2vec models) and \textit{contextual} word embeddings (such as transformers based models), on the task of patent similarity calculation. Second, it compares specifically the performance of Sentence Transformers (SBERT) architectures with different training phases on the patent similarity task. To assess the models' performance, we use information about \textit{patent interferences}, a phenomenon in which two or more patent claims belonging to different patent applications are proven to be overlapping by patent examiners. Therefore, we use these interferences cases as a proxy for maximum similarity between two patents, treating them as \textit{ground-truth} to evaluate the performance of the different embedding models. Our results point out that, first, \textit{Patent SBERT-adapt-ub}, the domain adaptation of the pretrained Sentence Transformer architecture proposed in this research, outperforms the current state-of-the-art in patent similarity. Second, they show that, in some cases, large static models performances are still comparable to contextual ones when trained on extensive data; thus, we believe that the superiority in the performance of contextual embeddings may not be related to the actual architecture but rather to the way the training phase is performed.
\end{abstract}

\section{Introduction}\label{sec:intro}

Patents provide their owners with a legal right to exclude others from making, using, selling, offering for sale, or importing the patented invention. This legal right is embodied in semi-structured text documents that describe the background, inventive concept, as well as the “metes and bounds” of the invention. 
\par In recent times, a growing body of research started to link patent analytics to natural language processing (NLP). This is possible because patents have textual attributes which constitute the patent itself: a title, an abstract, one or more claims and a description. In the last five years, the number of studies on information retrieval using patent text had a steep increase, covering different NLP tasks, such as name entity recognition (see for instance, the very recent study by \citet{puccetti2023technology} on how to extract technological entities from patent text), technological class classification \citep{tran2017supervised, lee2019patentbert, li2018deeppatent, roudsari2021comparison} and textual similarity calculation \citep{arts2021natural, bekamiri2021patentsberta, hain2022text}.

\par Textual similarity among patent (p2p) or patent portfolios is an essential tool to map, understand and predict the innovation pattern and dynamics of the actors involved in the innovation ecosystem. Recently, researchers have been using patent vector space models based on different NLP embeddings models to calculate the similarity between pairs of patents to help better understanding innovations, patent landscaping, technology mapping, and patent quality evaluation \citep{bekamiri2022survey}. 
Recently, replacing static embeddings with contextualized representations has yielded significant improvements on a diverse array of NLP tasks, ranging from question answering to coreference resolution. In the patent domain, on the one hand, authors previously underlined the fact that polysemy is especially significant when looking at patent data, because patent text is characterized by specific, technical, legal and therefore ambiguous jargon \citep{bonino2010review, tseng2007text}. Furthermore, patent text contains novel technologies by definition such that these vocabulary based methods would need to be retrained to include them in their dictionary. 
However, static embeddings have advantages in many scenarios, such as a much lower training cost \citep{strubell2019energy} and an easier interpretability \citep{gupta2021obtaining}. In recent works, both static \citep{whalen2020patent, hain2022text} and contextual \citep{bekamiri2021patentsberta} are used to measure patent similarity. Further, scholars underline the lack of a ground-truth benchmark dataset to compare and validate those similarity measures \citep{whalen2020patent, hain2022text, bekamiri2022survey} to establish the better approach.
\par Following the idea that for evaluating domain-specific word embeddings there is the need of a ground-truth dataset \citep{betti2020expert}, this research brings two original contributions on the textual patent similarity task. For both contributions, we use of a dataset about \textit{patent interferences}, which measure (nearly) simultaneous instances of identical invention by two or more independent parties: by recording instances of common invention, patent interferences create a unique record of common knowledge inputs \citep{ganguli2020paper}. In particular, the claims of patents involved in an interference can be considered as a case of maximum similarity between patent texts, therefore they can be used as a ground-truth benchmark dataset.  
First, using interferences data, we are able to assess the performance of the most relevant patent-specific deep learning models on the task of p2p similarity, testing whether pretrained \textit{static} embeddings or \textit{contextual} embeddings perform better on the patent similarity task. Second, we compare the performance of three different contextualized models, specifically Sentence Transformers (SBERT) models, following the idea that, among transformers based architectures, SBERT vectorial representations are best suited for
sentence-level downstream tasks compared to those produced by the vanilla BERT model and its variants \citep{reimers2019sentence, nikolaev2023representation}. In particular, we compare the augmented SBERT proposed by \citet{bekamiri2021patentsberta} to  two original Sentence Transformer models developed in this research, which are trained using pairs of patents which share the same label (technological CPC class):
\begin{enumerate}
\item \textit{Patent sBERT-ub}: a finetuned model from a pretrained RoBERTa model \citep{liu2019roberta}
\item \textit{Patent sBERT-adapt-ub}: a domain adaptation of a pretrained SBERT \citep{reimers2019sentence}
\end{enumerate}
to verify which model leads to the best performance.
\par The results show that, on the task of patent similarity, the domain adaptation of Sentence Transformers (Patent sBERT-ub) has the best performance; however, a static embedding model results', the word2vec proposed by \citet{hain2022text}, are comparable with transformers based ones. Therefore, we find that there is not a clear superiority of contextual over static embeddings or viceversa in the patent similarity task. This might be explained by the fact that, despite patent specific models would surely benefit of the contextual embeddings advantages, such as a non-fixed dictionary and polysemy aware embeddings, an extensive training phase still impacts the results. Furthermore, when comparing Sentence Transformers' models, the proposed domain adaptation which exploits golden labeled data leads to the best performance.
\par \par The rest of the paper is structured as follows: Section \ref{sec:lit} presents the state-of-the-art in deep-learning derived patent embeddings, Section \ref{sec:data} presents the data used, Section \ref{sec:models} depicts the models and the experiments, Section \ref{sec:results} analyzes the results, while Section \ref{sec:conclusion} concludes, acknowledging the limitations and opening the path for future developments.

\section{Related work}\label{sec:lit}

The use of text attributes to calculate patent similarity started from simple and straightforward keywords-based approaches, using co-occurrence or some weighted versions of it, such as point-wise mutual information (PMI) or n-grams \citep{arts2021natural}. However, keyword based approaches suffer of a number of shortcomings including the lack of consideration of the context to define the semantic meaning of words or the assessment of type of co-occurrence (synonymy or antonymy) as well as the computation expense and the \textit{curse of dimensionality} due to the large and very sparse matrices. 
\par A first step towards solving this kind of problems involved the creation static word vectors, which have demonstrated to be able capture meaningful syntactic and semantic regularities, with an architecture defined as \textit{word2vec} \citep{mikolov2013distributed} which takes into account the window of words which surrounds the word of interest. Word2vec architectures link each token to a vector creating a lookup matrix where there is a correspondence one to one; however, this kind of model is unable to produce differentiated vectorial representations for words that have a different meaning in different contexts. This problem is especially significant when looking at patent data, because patent text is characterized by specific technical, legal and often new jargon, where a single technology might be associated to a variety of uses and a single use to a variety of technologies.
\par For these reasons, newer attempts of creating meaningful patent vectors to calculate patent similarity involve the use of more recent neural networks architectures which are able to produce \textit{contextual representations} for patent text, meaning that the vectors change according to the surrounding context.  
Table \ref{tab:lit} summarizes some of the most relevant attempts of creating vectorial representation for patent text using neural networks (NN). 
These works are different to many respects, both conceptual and technical. 
\par First of all, some of them use the vectorial representation of patent texts to perform a multilabel classification task \citep{li2018deeppatent, lee2019patentbert, bekamiri2021patentsberta, hain2022text, roudsari2021comparison}. To reduce the burden over examiners and taking into account the increasing complexity of technological classes, many researcher focused on this task \citep{chen2012three, tran2017supervised, grawe2017automated, hepburn2018universal, li2018deeppatent, lee2019patentbert, bekamiri2021patentsberta, roudsari2021comparison}. To accomplish the task, different authors  not only use different architectures but also different textual parts and different validation methods. For instance, \citet{li2018deeppatent} propose their neural model \textit{DeepPatent} which encodes titles and abstracts of around 2 millions of USPTO patents using skipgram and perform the classification using a convolutional neural network (CNN) approach. The following work of \citet{lee2019patentbert} uses DeepPatent as the baseline to benchmark; they propose a new model called \textit{PatentBERT} using both the same data as \citet{li2018deeppatent} and a larger sample containing  3’050’615 patents. Further, they first compare their performance to DeepPatent using title and abstract, and then they make an original contribution using claims. In patent classification, the recent work of \citet{roudsari2021comparison} proved that contextual embeddings perform better compared to the static ones, while in \citet{bekamiri2022survey}, the authors show that models fine-tuned on patent data, such as the word2vec TF-IDF model presented in \citet{hain2022text}, perform better than generic contextual models.
\par Other authors create patents' embeddings which are explicitly aimed at patent similarity calculation. \citet{whalen2020patent} follow this second approach. In particular, their originality lies in the choice of the text to be used (description and independent claims) and in the model, as they train a Doc2vec architecture. Being trained on USPTO granted patent data from 1976 to 2019, their 300-dimensions embeddings provide document-level representations for each patent text\footnote{Further, their model was computed using the Gensim Python library, using the distributed bag of words (DBOW) algorithm with 10 epochs.}. Their validation procedure consists in checking the correlation between the obtained measure of patent similarity, measured as cosine among vectors, and being in the same technological field (CPC) or between citing and citetd patents. However, this kind of measures might have some drawbacks. First, patents in the same patent class might cover different technologies which share a similar purpose; at the same time, often the same patent is attributed to more than one technological class, creating further confusion. Second, the similarity relationship between citing and citetd patent is an indeed debated topic in the economics of innovation literature which relevant scholars state might be biased \citep{bacchiocchi2010international}. 
\par The work of \citet{bekamiri2021patentsberta} and the one of \citet{hain2022text} combine the multi-label classification task with the patent similarity calculation. \citet{bekamiri2021patentsberta} use USPTO granted patent data from 2013 to 2017, for a total of 1’492’294 patents. Their choice involves the use of patent claims, in particular of the first claim only, as it is usually the fundamental claim from which the rest of the patent application is derived from. Their technique is also original, as they propose an augmented SBERT approach, following the paper of \citet{thakur2020augmented}. The model is created according to the following steps: first they fine-tune RoBERTa as a cross-encoder over a small STS benchmark dataset; then, they use the trained RoBERTa to label 1'143 patent claims; finally, they train SBERT on the labeled target dataset (a small STS + 3'432 pair of claims sentences). Their overall result for the classification is represented by a F1 score in the multilabel classification task with CPC codes of 66\%. 
Finally, \citet{hain2022text} use around 48 millions of patents in English language filed at the EPO in a time range from 1980 to 2017. They use patents' abstracts only to produce a word2vec based encoding which is then multiplied by a TF-IDF weighted bag-of-word representations of the abstracts. As a result, they obtain a 300-dimensional patent signature vector that they use for similarity calculations.
For the multi-label classification task, they report a F1 of 52\%. Further, to validate their similarity measures, they investigate the correlation between their generated p2p similarity and the existing observable measures commonly used to approximate technological similarity, such as being filed from the same inventor, being in the same technological class or citing each other. However, \citet{hain2022text} acknowledge the limitations of this validation approach, very similar to the one proposed by \citet{whalen2020patent}, and stress the lack of a ground-truth to evaluate patent similarity.

\begin{landscape}
\begin{table}[]
\centering
\caption{Most recent contributions related to creation of word embeddings for patent text using neural networks}
\label{tab:lit}
\resizebox{\columnwidth}{!}{%
\begin{tabular}{lllllll}
\hline
\textbf{Auhtor(s)} & \textbf{Year} & \textbf{Data}                                                                                                                  & \textbf{Text}                                                                           & \textbf{Task}                                                                                                         & \textbf{Technique}                                                                 & \textbf{\begin{tabular}[c]{@{}l@{}}Performance/\\ validation\end{tabular}}                                                                                                                                \\ \hline
Li et al.          & 2018          & \begin{tabular}[c]{@{}l@{}}USPTO 2006-2015 \\ (2'000'147)\end{tabular}                                                         & 1.title and abstract                                                                    & \begin{tabular}[c]{@{}l@{}}multi-label classification \\ (IPC codes)\end{tabular}                                     & \begin{tabular}[c]{@{}l@{}}skipgram + \\ CNN\end{tabular}                          & \begin{tabular}[c]{@{}l@{}}Precision: \\  73.88\%\\ \\ F1: \\ 43\% top 5\end{tabular}                                                                                                                     \\ \hline
Lee and Hsiang     & 2020          & \begin{tabular}[c]{@{}l@{}}USPTO (two versions: first as in Li et al.  \\ second of 3'050'615 obs. )\end{tabular}                            & \begin{tabular}[c]{@{}l@{}}1.title and abstract\\ 2.claims\end{tabular}                 & \begin{tabular}[c]{@{}l@{}}multi-label classification \\ (IPC and CPC codes)\end{tabular}                             & \begin{tabular}[c]{@{}l@{}}fine-tuning of\\ BERT uncased +\\ CNN\end{tabular}      & \begin{tabular}[c]{@{}l@{}}v1(IPC):\\ \\ Precision:\\ 30.31\%\\ F1:\\ 45\% top 5\\ \\ v2(CPC):\\ Precision best top 1:\\ 85\%\\ F1 best top 1:\\ 67\%\end{tabular}                                        \\ \hline
Whalen et al.      & 2020          & \begin{tabular}[c]{@{}l@{}}USPTO (utility) granted\\ (1976-2019)\footnote{the number of patents has not been specified in the paper}\end{tabular} & \begin{tabular}[c]{@{}l@{}}1. description\\ 2. independent claim(s)\\ text\end{tabular} & Patent similarity calculation                                                                                         & Doc2vec (trained)                                                                  & \begin{tabular}[c]{@{}l@{}}Correlation between patent similarity\\ and being in the same CPC field and\\ between citing and citetd patents (backward\\ and forward citations)\end{tabular}                 \\ \hline
Bekamiri et al.    & 2021          & \begin{tabular}[c]{@{}l@{}}USPTO 2013-2017 \\ (1'492'294)\end{tabular}                                                         & 1.first claim                                                                           & \begin{tabular}[c]{@{}l@{}}Patent similarity calculation and\\ multi-label classification (CPC\\ codes)\end{tabular}  & \begin{tabular}[c]{@{}l@{}}augmented approach for\\ fine-tuning SBERT\end{tabular} & \begin{tabular}[c]{@{}l@{}}F1:\\ \\ 66\%\end{tabular}                                                                                                                                                     \\ \hline
Hain et al.        & 2022          & \begin{tabular}[c]{@{}l@{}}EPO granted 1980-2017\\ (around 12 millions of patent applications)\end{tabular}                    & 1.abstract                                                                              & \begin{tabular}[c]{@{}l@{}}Patent similarity calculation and \\ multi-label classification (IPC\\ codes)\end{tabular} & \begin{tabular}[c]{@{}l@{}}Word2vec TF-IDF weighted \\ BoW\end{tabular}          & \begin{tabular}[c]{@{}l@{}}F1:\\ \\ 52\%\\ \\ Correlation between being in the same IPC\\ class and being similar and between citetd \\ and citing patents (backward and forward\\ citations)\end{tabular} \\ \hline
\end{tabular}%
}
\end{table}
\end{landscape}

\section{Data} \label{sec:data}

Two datasets are crafted in this paper and both are created using online free available data only. The former dataset includes the necessary triplets for creating the two SBERT derived models using data from Patents View; the latter is the \textit{patent interference} dataset, which, after some preprocessing operations explained in Section \ref{subsec:inter}, will be used as ground-truth benchmark dataset for evaluating the different similarities measures, as it reports some couple of claims which experts at the USPTO judged as covering the exact same invention. More detail about both datasets is provided in the following subsections.

\subsection{Triplets creation: anchor, positive and negative} \label{sec:triplets}

The first dataset used in this paper is the one including the necessary triplets for creating the two proposed SBERT versions. In particular, the triplets are created starting from three tables from Patents View. The Patents View platform is built on a regularly updated database that longitudinally links inventors, their organizations, locations, and patent characteristics. The data collected from Patents View include three kinds of information which are useful for creating the triplets\footnote{The data is free and accessible at \url{https://patentsview.org/download/data-download-tables}}:
\begin{enumerate}
    \item Information on the technological CPC class(es) assigned to each patent application (Table \textit{g\_cpc\_current})
    \item Information on the filing date of the patent (Table \textit{g\_application})
    \item Patents' abstracts (Table \textit{g\_patent})
\end{enumerate}

In particular, the Table \textit{g\_cpc\_current} contains information on 7'469'689 patents, among which we select 940'564 that are assigned only to one CPC code. We do this as our goal is to pair similar patents and the easiest choice is to pair those having the same CPC code. Further, we select only those patents whose CPC code is defined as "inventional", considering the other type, the "additional", usually identifies additional classes to which the patent can be assigned and not the main one.
The second step is to attach to each patent the information about the filing date, coming from the Table \textit{g\_application}. Third, we retrieve the abstract of each patent using the Table \textit{g\_patent}. At this point we select only utility patents, for a total of 939'517 observations for which we have patent identifier, filing date, the abstract and the single CPC class to which it is assigned. For the purpose of this research, we consider two patents belonging to the same CPC class if their CPC section, CPC class, CPC group and CPC subgroup are the same. After, we create groups of patents which share the same CPC class and filing year and delete those groups which do not contain two patents at least, cutting our observations to 568'086. We select patent pairs filed close in time as patenting activities are usually consistent with technological trends. In this sense, patents filed close in time might be more similar than those filed years apart. Further, considering that technological trends might drastically change in very short periods of times \citep{sobolieva2021global} 
\footnote{For instance, the history of Kodak is an example of such dramatic changes. In 1998, the company produced 85\% of
the world's volume of paper for cameras and there were 170 thousands employees. In fact, three years later digital cameras have replaced the company with its market and it became bankrupt. However, different kinds of technologies might have technological cycles of different length.}, we select couples of technologies filed in the same civil year.
After, doing all possible pairwise combinations among patents which have the same CPC code and are filed in the same year, we obtain 1'537'809 pairs; in addition, we exclude those couples whose abstract is the same.\footnote{The phenomenon according to which two different patents have the same abstract is known as \textit{patent continuation}, where a new patent is filed by an applicant who wants to pursue additional claims to an invention disclosed in an earlier application. More information on this issue could be found, among other, in \citet{righi2020patenting}.}
For each of the created pairs (anchor and positive), we randomly add an abstract (negative) from a patent which does not belong the same CPC class of the two others in the pair.

\subsection{Patent interferences: a dataset for validation} \label{subsec:inter}

Section \ref{sec:intro} quickly introduced the phenomenon of patent interferences. To recall, an \textit{interference} takes place when an examiner at the patent office finds two applications whose claims are partially or totally overlapping and therefore a panel of judges is needed to decide who put together the idea first. The recent work by \citet{ganguli2020paper} explores this issue and provides a public dataset which contains information on 1'324 interferences cases, filed from 1998 to 2014, involving USPTO patents\footnote{The dataset is available in the "Additional material" section at the following url \url{https://www.aeaweb.org/articles?id=10.1257/app.20180017}}. For each interference, they provide the number of patent applications which are involved in the cases. However, there is no information on which the exact claims involved from each application are and the claims themselves are not provided along with the interferences dataset. For these reasons, we complement the information in the interferences dataset with data from Patents View and we create a strategy to select the most likely couple of claims object of the interference. In particular, for each patent application we first retrieve all its pre-grant claims.\footnote{Patent claims information, divided by year, is available at \url{https://patentsview.org/download/pg_claims}} Pre-grant claims information is available only from 2001 onwards, therefore we have to limit our dataset to interferences cases which involved patents filed between 2001 and 2014 included, leaving us with 219 interferences cases linked to 338 patent applications. Furthermore, we exclude those interferences where we do not have information about the claims for both the patents involved and those including more than two patents at a time; in that way, we obtain a dataset which contains 133 interferences linked to 260 unique patent applications.\footnote{The same patent application can be involved in more than one interference.} After this step, for each of the patent applications we consider the independent claims\footnote{Usually patents have a mixture of independent and dependent claims; the main difference among the two is that, while the former can "stand alone", the latter are linked to the former for both technological and semantic content.} only, for a total, after performing some preprocessing operations.\footnote{For example, we deleted those claims which included the word "canceled/cancelled" as they were not informative about the content of the patent application.},  of 1'447 total unique independent claims. With those claims, we generate all the possible combinations belonging to two distinct patent applications in the same interference, obtaining a total of 5'537 couples. Among those, for each interference we select the couple with the greatest patent similarity, measured using the encoding of SBERT \citep{reimers2019sentence}. At the end, we retain 133 pairs of claims, one for each of the interference cases.
Table \ref{tab:tabclaim} reports some of the couples that were selected as an example. The interference data in this form will be used as ground-truth benchmark dataset to evaluate the performance of the different models proposed in literature to estimate patent similarity and compare them with those originally proposed in this research.\footnote{It would be useful to compare this dataset to the one used for fine tuning, as presented in Section \ref{sec:triplets}. However, the dataset presented in this section only has 133 pairs of sentences which we then use for validation of the different models; therefore, an analysis comparing the two would not be helpful. This is a limitation of this research which is further discussed in Section \ref{sec:conclusion}.}

\clearpage
\begin{landscape}
\begin{table}[ht]
\tablefont
\caption{Example of patent interferences claims' couples}
\label{tab:tabclaim}
\begin{tabular}{p{0.6in}p{4in}p{4in}}
\hline
\textbf{No.} &
  \textbf{Claim 1} &
  \textbf{Claim 2} \\ \hline
\textbf{105075} &
  A method for the preparation of citalopram wherein the aldehyde of formula 16 is converted to the corresponding 5-cyano compound of formula (i) 17 which is alkylated to form citalopram, which is isolated in the form of the base or an acid addition salt thereof &
  A production method of citalopram represented by the formula which comprises reacting a compound of the formula with 3-(dimethylamino) propyl chloride in the presence of a condensing agent and at least one member selected from tetramethylethylenediamine and 1,3-dimethyl-2-imidazolidinone. \\ \hline
\textbf{105099} &
  An antibody that binds to mouse flt-3 ligand, wherein said mouse flt-3 ligand comprises amino acids 28-163. &
  An antibody that is immunoreactive with a flt3-l polypeptide. \\ \hline
\textbf{105224} &
  A glass door for a refrigerated display case, the door comprising: a first glass panel having in inside and an outside surface.  A low emissivity coating on the inside surface of the first glass panel; a second glass panel having in inside and an outside surface; a low emissivity coating on the inside surface of the second glass panel; an intermediate glass panel between the first and second glass panels; a first spacer assembly between the first and intermediate glass panels and a second spacer assembly  between the intermediate and second glass panels wherein the first and second spacer assemblies are formed from warm edge spacer assemblies; and a frame extending about and supporting at least one of the glass panels. &
  A refrigeration door having an outer surface and adapted to be mounted on a refrigerating compartment, said door comprising: a first sheet of glass;  a second sheet of glass; a first sealant assembly disposed around the periphery of said first sheet of glass and said second sheet of glass for maintaining said first sheet and said second sheet in spaced-apart relationship from each other; a first low emissivity coating adjacent a surface of said first sheet or said second sheet of glass;  said first sheet and second sheets of glass, said first sealant assembly, and said first low emissivity coating forming an insulating glass unit having a u value substantially equal to or less than 0.2 btu/hr-sq ft-f; and a frame secured around the periphery of said insulating glass unit. \\ \hline
\textbf{105236} &
  A juvenile seat assembly is provided for use with both a vehicle seat and anchor mounts provided in or near the vehicle seat, the assembly comprising a juvenile seat having a first pair of openings and a second pair of openings, an anchor belt including a strap having a central portion and opposite end portions, and a connector coupled to each end portion and adapted to be coupled one of the anchor mounts, the anchor belt being threaded through the first openings to position the seat in a first, rearwardly facing position and the anchor belt being threaded through the second openings to position the seat in a second, forwardly facing position, and a leash coupled to the juvenile seat and to the central portion of the anchor belt, the leash remaining coupled to the seat and to the anchor belt when the anchor belt is moved between the first and second openings. &
  An improved child car seat having a plurality of belt restraining guides for securing the child car seat to an automobile seat, the improvement comprising: a. an installation belt having a belt length adjuster, a belt first end having a fastener, a belt second end having a fastener, and a belt intermediate region between the ends, and; b. a tether having a tether first end attached to an anchor point on the child car seat and a tether second end attached to an anchor point on the installation belt, wherein the installation belt is anchored to the child car seat only by the tether. \\ \hline
\textbf{105252} &
  An implantable device for insertion into a cavity in a vertebral body comprising: a flexible container having a wall membrane: said wall membrane defining an interior and an exterior of said container; said wall having at least one hole connecting the interior with the exterior; a fill tube coupled to said container at a location proximate an edge of said container for injecting a flowable or fluid bone filler material into said container such that said bone filler leaves said interior and enters said vertebral body. &
  An implantable device for insertion into a cavity in a vertebral body comprising: a container including; an upper wall member; a lower wall member; a circumferential wall  member; said wall members together defining a single chamber. \\ \hline
\end{tabular}
\end{table}
\scriptsize
\emph{Notes:} The table reports 5 examples of couples of claims belonging to the two patent applications involved in a certain interference; column 1 reports the interference number, while column 2 and column 3 reports the couple of most similar claims from the two patent applications.
\end{landscape}
\clearpage

\section{Models and experiments}\label{sec:models}
This research first tests whether static or contextual embeddings perform better on the patent similarity task. In particular, we consider the following models from the literature:
\begin{itemize}
    \item Word2vec TF-IDF model by \citet{hain2022text}
    \item Doc2vec model by \citet{whalen2020patent}
    \item Augmented SBERT model (Patent sBERTa) by \citet{bekamiri2021patentsberta}
\end{itemize}
\par Furthermore, this research proposes two original models to be compared with the others, both deriving from the Sentence Transformers architecture. To fine-tune both architectures, we exploit the triplets dataset as explained in Section \ref{sec:triplets}.
:
\begin{enumerate}
    \item \textbf{Patent SBERT-ub}: this model is a fine-tuning of the RoBERTa base model by \citet{liu2019roberta} which uses a random 10\% of our triplets dataset\footnote{The RoBERTa model used is available at \url{https://huggingface.co/roberta-base}, we trained this model with a batch size of 8 and 1 epoch.}
    \item \textbf{Patent SBERT-adapt-ub}: this model is a domain adaptation of the SBERT model presented in \citet{reimers2019sentence} which uses a random 10\% of our triplets dataset \footnote{\label{foot:1}The SBERT model used is available at \url{https://huggingface.co/sentence-transformers/all-MiniLM-L6-v2}, we trained this model with a batch size of 8 and 1 epoch. }
\end{enumerate}

Those two models have been trained using the architecture proposed by \citet{reimers2019sentence}. The dataset described in Section \ref{sec:triplets} has been split in traning and validation using respectively 70\% and 30 \% of the data. Further, in both cases we used mean as pooling strategy. As for the loss function, we used the \textit{TripletLoss} function\footnote{ For further detail please refer to the sentence-transformers package documentation \url{https://www.sbert.net/docs/package_reference/losses.html}.} introduced by \citet{schroff2015facenet}, where the distance is calculated using euclidean distance. The minimum margin for the distance between the anchor and the negative has been set to 5. Finally, we used the validation step each 1'000 training steps and in both cases we used 1 epoch for the training.

\par Second we test, among transformers based models, which architecture performs best in the patent similarity task; therefore, for the second analysis, we limit the analysis to Patent sBERTa, Patent SBERT-ub and Patent SBERT-adapt-ub.

\par To evaluate our results,  we exploit the dataset of \citet{ganguli2020paper} about patent interferences, whose preparation steps are explained in the Section \ref{sec:data}. In particular, knowing that the pairs of claims in the interference shall represent a case of maximum similarity, we define the best performing model the one which predicts in the majority of the cases the highest cosine similarity among those sentences. Furthermore, to prevent the case in which a certain model might overestimate the closeness among claims in general, we compare the results of the maximum similarity with those of the same dataset in which we create random couples of claims belonging to different interferences. In this case, we define the best performing model as the one which predicts the lower cosine similarity among the randomly generated pairs in most of the cases.

\section{Analysis} \label{sec:results}

Table \ref{tab:res1} reports the results of the comparison among the five models described in Section \ref{sec:models} on the task of patent similarity on the interferences dataset, veryfing whether static or contextual embeddings perform better on this specific task. The first column reports the name of the considered model, the second column the percentage of cases in which it reported the highest cosine similarity for each interference and the third one the percentage of cases in which a certain model reported the minimum value for cosine similarity among randomly generated pairs of claims. The model which has the greatest figures in both cases is Patent-SBERT-ub-adapt. Interestingly, the word2vec TF-IDF model by \citet{hain2022text} performs better than the doc2vec by \citet{whalen2020patent} and the SBERT by \citet{bekamiri2021patentsberta}. This might be due to the fact that the word2vec model has been trained on a very large number of abstracts (48 millions), therefore the representation for each patent abstract is quite accurate. On the other hand, Patent SBERTa has been fine tuned using 3'432 pairs of claims only. This highlights that static embeddings could also have good performances on specific tasks when their training is large enough even when comparing with more recent architectures such as transformers.

\begin{table}[]
\centering
\caption{Percentage of cases of greatest and lowest similarity by model (all)} 
\label{tab:res1}
\resizebox{\columnwidth}{!}{%
\begin{tabular}{lll}
\hline
\textbf{Model} &
  \textbf{\begin{tabular}[c]{@{}l@{}}Max similarity \\ (\%)\end{tabular}} &
  \textbf{\begin{tabular}[c]{@{}l@{}}Min similarity \\ (\%)\end{tabular}} \\ \hline
\textit{Patent SBERT\_ub\_adapt} & 52 & 40 \\ \hline
\textit{Patent SBERT\_ub}        & 32 & 26 \\ \hline
\textit{Word2vec TF-IDF} & 11 & 15\\ \hline
\textit{Doc2vec}                & 4  & 6 \\ \hline
\textit{Patent SBERTa}          & 1  & 13 \\ \hline
\end{tabular}%
}
\scriptsize
\emph{Notes:} The table reports the results of the comparison among the five models (Word2vec TF-IDF, Doc2vec and Patent SBERTa from previous works plus the two original models). In particular, the second column shows the no. of times, in percentage, in which each of the models assigned the maximum similarity score to each of the 133 couples of the interferences dataset; the third column reports in fact the no. of  times, in percentage, in which each of the models assigned the lowest similarity score for the randomly generated pairs of patent claims belonging to the interferences dataset.

\end{table}

\begin{table}[] 
\centering
\caption{Percentage of cases of greatest and lowest similarity by model (SBERT only)}
\label{tab:res2}
\resizebox{\columnwidth}{!}{%
\begin{tabular}{lll}
\hline
\textbf{Model} &
  \textbf{\begin{tabular}[c]{@{}l@{}}Percentage with\\ max similarity\end{tabular}} &
  \textbf{\begin{tabular}[c]{@{}l@{}}Percentage with\\ min similarity\end{tabular}} \\ \hline
\textit{Patent SBERT\_ub\_adapt} & 60 & 47 \\ \hline
\textit{Patent SBERT\_ub}        & 37 & 30 \\ \hline
\textit{Patent SBERTa}          & 3  & 23 \\ \hline
\end{tabular}%
}
\scriptsize
\emph{Notes:} The table reports the results of the comparison among the three SBERT models (Patent SBERTa from \citet{bekamiri2021patentsberta} and 2 original). In particular, the second column shows the no. of times, in percentage, in which each of the models assigned the maximum similarity score to each of the 133 couples of the interferences dataset; the third column reports in fact the no. of times, in percentage, in which each of the models assigned the lowest similarity score for the randomly generated pairs of patent claims belonging to the interferences dataset.
\end{table}

\par Table \ref{tab:res2} reports the results for the comparison among SBERT models only. From the table, we acknowledge that, among the three SBERT architectures, the best performing one is the domain adaptation of the SBERT by \citet{reimers2019sentence}. The better performance of the domain adaptation of SBERT is consistent with the idea that it is valuable to complement work on large LMs  and use domain and task relevant corpora to specialize models \citep{gururangan2020don}.

\section{Conclusion} \label{sec:conclusion}

This paper enriched the discussion about textual patent similarity in several ways: first of all, it analyzed the differences among performance of pretrained static versus contextual embeddings on the task of patent similarity and proposed a new SOTA contextual model (Patent-SBERT-ub-adapt); second, it compared different architectures of SBERT models to identify which one better performs  for the patent similarity task. Consistently, from the results we can derive two main contributions: first of all, a contextual model is the best performer on the patent similarity task; however, we demonstrated that not always the more recent contextual embeddings perform better than static embeddings as well. Therefore, we claim that the performance is also linked to the task itself and to the ways in which the training of the models took place. Second, among SBERT architectures, Patent-SBERT-ub-adapt is the one which leads to the best results, confirming the benefits of domain adaptation for pretrained models \citep{gururangan2020don} .
\par However, this research is not free from limitations. For instance, the domain adaptation of SBERT might be less general than the original architecture by \citet{reimers2019sentence}. Another important limitation is related to the fact that we do not have a ground-truth labeled dataset for very dissimilar patents, therefore we just create random couples of claims. The best option would be in fact to have another manually labeled dataset. At the same time, a bigger ground-truth dataset could improve the overall reliability of this analysis and allow allineation between the dataset using for fine-tuning and the validation dataset.
\par Eventually, future work could improve the current creation process of the triplets dataset. In particular, the triplets could be created using not only the patents' asbtracts but also one or more claims and exploit their hierarchy as well.

\section*{Acknowledgments}

This research was funded by CNRS as part of the 2022-2023 "Programme Prématuration".
\clearpage

\bibliographystyle{acl_natbib}
\bibliography{references.bib}

\end{document}